\def\year{2022}\relax
\documentclass[letterpaper]{article} 
\usepackage{aaai22}  
\usepackage{times}  
\usepackage{helvet}  
\usepackage{courier}  
\usepackage[hyphens]{url}  
\usepackage{graphicx} 
\usepackage{natbib}  
\usepackage{caption} 
\DeclareCaptionStyle{ruled}{labelfont=normalfont,labelsep=colon,strut=off} 
\usepackage{algorithm}
\usepackage{algorithmic}
\usepackage{subcaption}
\usepackage{amsmath}
\usepackage{amssymb}
\usepackage{amsfonts}
\usepackage{stackrel}   
\usepackage{multirow}
\usepackage{color}
\usepackage{comment}
\usepackage{flushend}

\usepackage{newfloat}
\usepackage{listings}
\floatstyle{ruled}
\newfloat{listing}{tb}{lst}{}
\floatname{listing}{Listing}
\title{TA$^2$N: Two-Stage Action Alignment Network for Few-Shot Action Recognition}
\author{
Shuyuan Li{\normalsize$^{1}$}\footnote{Equal contribution.}, Huabin Liu{\normalsize$^{1}$}\footnotemark[\value{footnote}], 
Rui Qian{\normalsize$^{1}$}, Yuxi Li{\normalsize$^{1}$}, John See{\normalsize$^{2}$} \\ Mengjuan Fei{\normalsize$^{3}$}, Xiaoyuan Yu{\normalsize$^{3}$}, Weiyao Lin{\normalsize$^{1}$}\footnote{Corresponding author.}
}

\affiliations{
    \textsuperscript{\rm 1} Shanghai Jiao Tong University, Shanghai, China \\
    \textsuperscript{\rm 2} Heriot-Watt University Malaysia, Putrajaya, Malaysia\\
    \textsuperscript{\rm 3} Huawei Cloud, China\\

    \{shuyuanli,huabinliu,qrui9911,lyxok1\}@sjtu.edu.cn, J.See@hw.ac.uk \\
    \{feimengjuan1,yuxiaoyuan\}@huawei.com,  wylin@sjtu.edu.cn
}

\usepackage{bibentry}

\begin{document}

\maketitle

\begin{abstract}
Few-shot action recognition aims to recognize novel action classes (query) using just a few samples (support). The majority of current approaches follow the metric learning paradigm, which learns to compare the similarity between videos. Recently, it has been observed that directly measuring this similarity is not ideal since different action instances may show distinctive temporal distribution, resulting in severe misalignment issues across query and support videos. In this paper, we arrest this problem from two distinct aspects -- action duration misalignment and action evolution misalignment. We address them sequentially through a Two-stage Action Alignment Network (TA$^2$N). The first stage locates the action by learning a temporal affine transform, which warps each video feature to its action duration while dismissing the action-irrelevant feature (e.g. background). Next, the second stage coordinates query feature to match the spatial-temporal action evolution of support by performing temporally rearrange and spatially offset prediction. Extensive experiments on benchmark datasets show the potential of the proposed method in achieving state-of-the-art performance for few-shot action recognition. The code of this project can be found at \textcolor{magenta}{https://github.com/R00Kie-Liu/TA2N}.
\end{abstract}

\section{Introduction}

\label{sec:intro}
Action recognition~\cite{c3d2015,I3d2017,TSN2016} has received considerable attention in the computer vision community due to the increasing demand for video analysis in real-world scenarios. In recent years, deep learning methods have dominated the field of video action recognition with convolutional neural networks (CNNs). Numerous labeled data empower these CNNs-based methods to train a discriminative classifier for a finite set of classes. Nevertheless, in a real sense, the number of novel action categories may be limited. This problem is compounded by the laborious and expensive task of annotating all available videos for these novel categories. 
\begin{figure}[t]
    \begin{center}
        \includegraphics[width=0.9\linewidth]{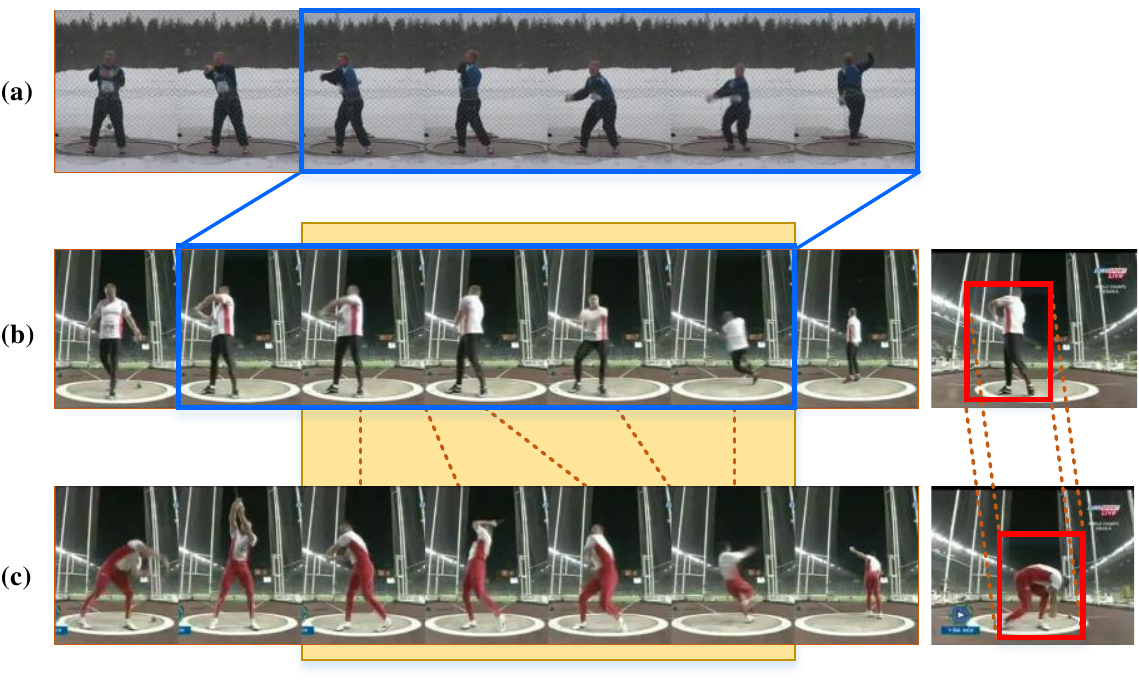}
    \end{center}
    \caption{Example of action misalignment. (a)\&(b): \textit{action duration misalignment}. The action duration  is highlighted with blue rectangles. (b)\&(c): \textit{action evolution misalignment} in temporal (left) and spatial (right) aspects. The red dashed connected lines indicate pairs of temporal or spatial positions are consistent in action content. The action category of these presented videos is \textit{`Hammer throw'}.}
    
    \label{fig:misalign}
\end{figure}
Consequently, the few-shot learning (FSL) task, which aims to recognize novel visual categories from very few labeled examples, has come into prominence in recent years. The solutions for image-based few-shot learning fall into three general categories: metric learning~\cite{prototypical2017,relationNet2018,matching2016}, data augmentation~\cite{fsl-augment1, fsl-augment2}, and optimization-based methods~\cite{FSL-optimization-1,FSL-optimization-2}. Each of them has made impressive progress in general image recognition. However, fewer studies have been carried out on few-shot video action recognition.
The majority of existing approaches in this area follow the metric learning-based paradigm~\cite{compound2018, taen2020, tarn2019}, which learns to compare the similarity between the videos from known classes and videos from novel classes. Recently, some research works~\cite{tarn2019,ARN2020,otam2020} observed that it is challenging to directly measure the similarity between videos due to the fact that different action instances show distinctive temporal distributions, \emph{e.g.}, different temporal locations or evolution processes, along the timeline in videos, can result in severe misalignment issues between the query and support videos. Some methods attempted to address this by performing temporal alignment, \emph{e.g.}, TARN~\cite{tarn2019} proposed a segment-by-segment attention module to perform temporal alignment at feature level; 
ARN~\cite{ARN2020} designed attention mechanisms to locate the discriminative temporal blocks.
In contrast to these works, OTAM~\cite{otam2020} explicitly aligns video sequences with a variant of the dynamic time warping algorithm. Aligning the semantic content in videos is still challenging since there exists wide variation of action instance. As such, the problem of video alignment in few-shot action recognition remains quite under-explored. 

In this paper, we delve into this specific problem in few-shot action recognition from two distinct aspects -- both indicative of distinct misalignment issues. First, the relative temporal location of an action is usually inconsistent between videos due to different start time and duration (as shown in Fig.~\ref{fig:misalign}(a)\&(b)); in this paper, we define the issue of location inconsistency as \textbf{\textit{action duration misalignment}} (ADM).
Second, since action often evolves in a non-linear manner,  the discriminative temporal-spatial part within the action process varies from action instances (as shown in Fig.~\ref{fig:misalign}(b)\&(c), left and right respectively), even though they share the same semantic category and duration. We define this internal spatial-temporal malposition among action instances as \textbf{\textit{action evolution misalignment}} (AEM).

To cope with these two types of misalignment, we devise a Two-stage Action Alignment Network (TA$^2$N) for few-shot action recognition. The first stage utilizes a Temporal Transformation Module (TTM) to predict temporal warp parameters for input video and then perform an affine transformation on the feature sequence, aligning it with the action duration period. In the second step, an Action Coordinate Module (ACM) is adopted to align the action evolution from both temporal and spatial aspects. For temporal evolution, similar motion patterns across action processes should be aggregated into the same temporal location. Thus, in temporal coordination (TC), we model the motion correlation between paired support and query videos, and then the query features could be temporally rearranged to match the support one according to the highly correlated positions. As for spatial coordination (SC), the action-specific regions (\emph{e.g.} actors) of paired frames are required to be spatially consistent in position. Accordingly, we predict spatial offset for paired frames and then perform the corresponding movement over spatial features to align them. Benefit from our proposed two-step and coarse-to-fine align strategy, actions could be well aligned in duration and evolution, advocating a more accurate metric learning and classification.
The detailed pipeline of our proposed method is illustrated in Fig.~\ref{fig:framework}. 

In summary, our main contributions are as follows:
\begin{itemize}
    \item We delve specifically into the misalignment problem in few-shot action recognition, revealing and quantifying two critical aspects of this issue: the action duration and evolution misalignment.
    \item We propose a novel two-stage action alignment network (TA$^2$N), which performs a jointly spatial-temporal action alignment over videos, to address these two aspects of misalignment sequentially.
    \item Extensive experiments show that our proposed method could relieve the misalignment and achieve state-of-the-art results in few-shot video action recognition.
\end{itemize}

\section{Related Work}
\textbf{Few-shot Learning} 
A primary challenge faced in FSL is the insufficiency of data in novel classes. The direct approach to address this is to enlarge the sample size by data augmentation. Some approaches~\cite{fsl-augment1,fsl-augment2} were proposed to generate unseen data with labels to enrich the feature spaces of novel classes. Autoaugment~\cite{autoaugment} further automatically learns the augmentation policy to improve the generalization on various few-shot datasets. Besides, learning metrics to compare the seen and novel classes is another popular way of handling FSL. Matching network~\cite{matching2016} is an end-to-end trainable kNN model using cosine as the metric, with an attention mechanism over a learned embedding of the labeled samples to predict the categories of the unlabeled data. The Prototypical Network~\cite{prototypical2017} uses a feed-forward neural network to embed the task examples and perform nearest neighbor classification with the class centroids. Relation Net~\cite{relationNet2018} proposed a novel network which concatenates the feature maps of two images, and proceeds to send the concatenated features to a relation net to learn the similarity. While these methods perform well on image recognition tasks, it is less optimal
to transfer them directly to action recognition.

\noindent
\textbf{Action Recognition} 
The state-of-the-art action recognition methods are focused on designing architectures with temporal modeling in mind. C3D~\cite{c3d2015} and I3D~\cite{I3d2017} are the most representative networks that extend VGGNet and InceptionNet respectively to 3D versions for extracting temporal information from videos. However, they lead to expensive computational costs and memory demand. Therefore, recent research has paid more attention towards efficient models such as P3D~\cite{p3d2017} and R(2+1)D~\cite{r(2+1)d2018}. These models decompose the 3D convolution into a 2D convolution and a 1D convolution to learn the spatial and temporal information separately.

\noindent
\textbf{Few-shot Action Recognition} 
Early studies on few-shot action recognition could be traced back to CMN~\cite{compound2018}, which proposed a compound memory network to store matrix representations and features can be easily retrieved and updated in an efficient way.
The majority of current studies on few-shot action recognition follow the metric learning paradigm. TAEN~\cite{taen2020} encodes actions in videos as trajectories in a metric space by a collection of temporally ordered sub-actions, whereby   FAN~\cite{dynamiciamge2019} then condenses the video motion feature into a single dynamic image, which relieves the pressure of learning the distance metrics. Due to the various temporal location of actions in videos, directly comparing the similarity of two videos with misaligned actions may lead to a sub-optimal distance metric. To solve this issue, some approaches proposed to perform temporal alignment. TARN~\cite{tarn2019} proposed an attentive relation network to perform the temporal alignment implicitly at the video segment level. OTAM~\cite{otam2020} explicitly aligns video sequences with a variant of the Dynamic Time Warping (DTW) algorithm. ARN~\cite{ARN2020} generates attention masks to re-weight spatiotemporal features. It utilizes augmentation strategies with self-supervised learning to enhance its feature encoder and attention mechanism. Among these methods, OTAM is the most related to our work.

\section{Quantifying Temporal Misalignment}
\begin{figure}
    \centering
    \begin{minipage}{0.32\linewidth}
    \includegraphics[width=1\textwidth,height=0.9\textwidth]{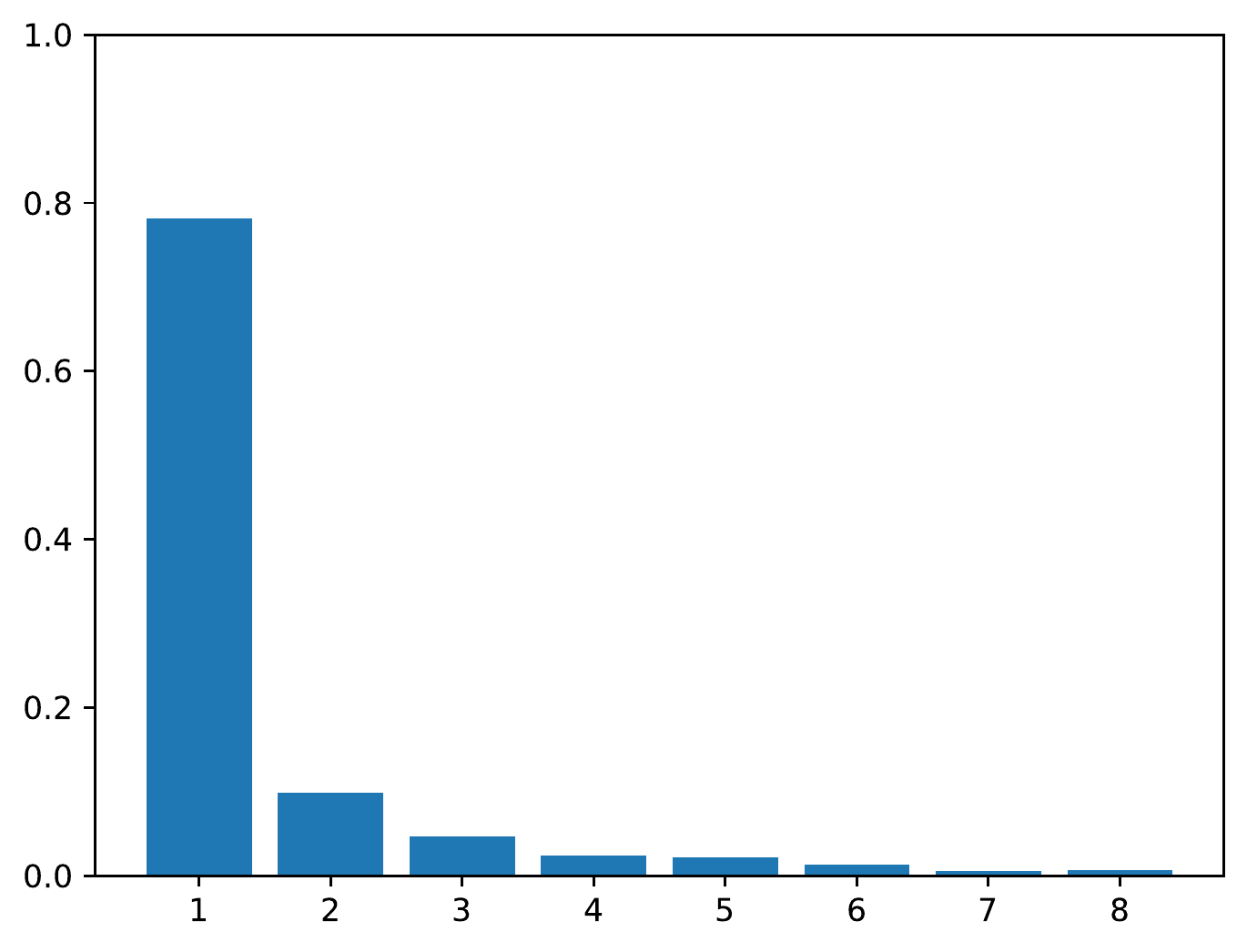}
    \subcaption{UCF101}
    \end{minipage}
    \begin{minipage}{0.32\linewidth}
    \includegraphics[width=1\textwidth, height=0.9\textwidth]{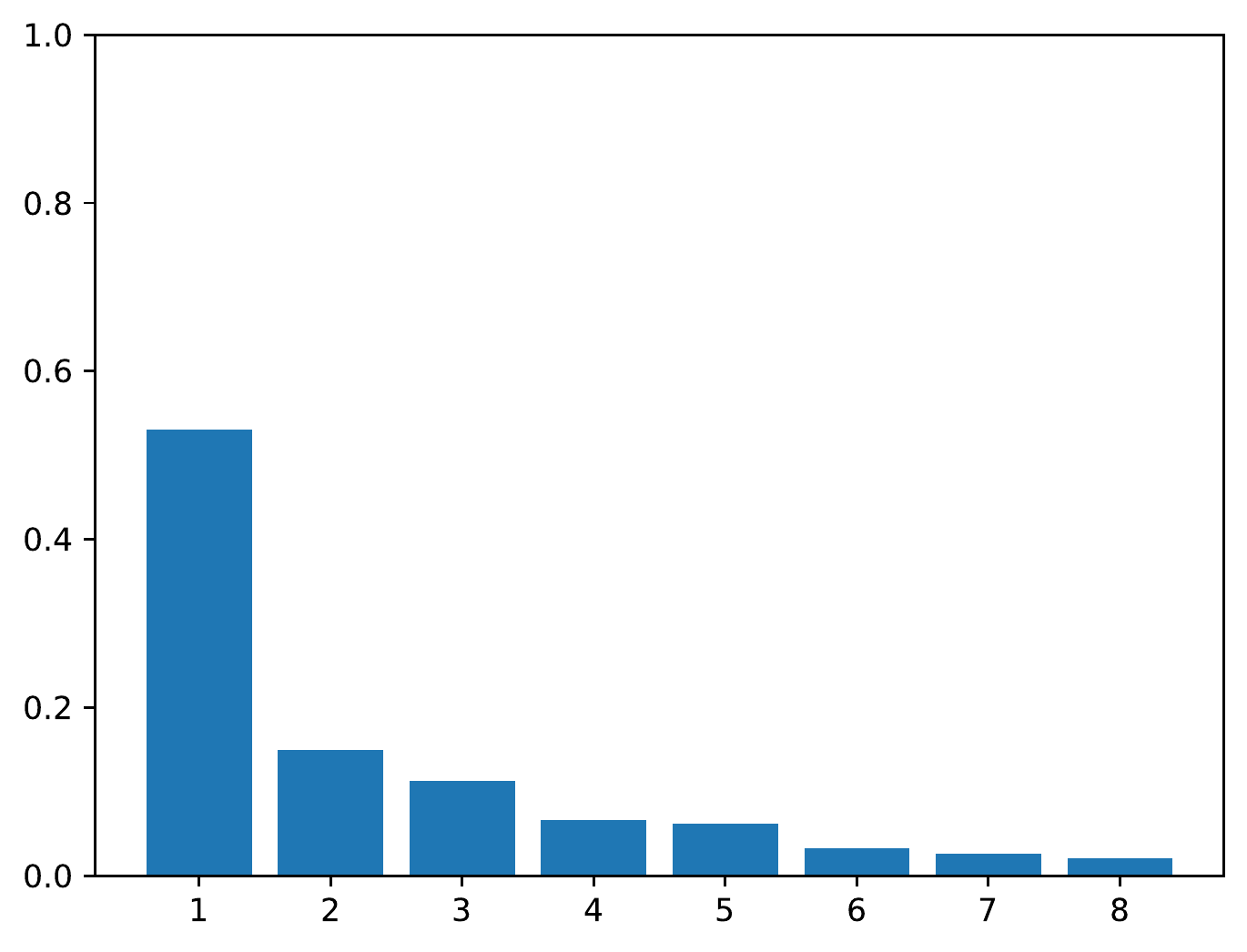}
    \subcaption{HMDB51}
    \end{minipage}
    \begin{minipage}{0.32\linewidth}
    \includegraphics[width=1\textwidth,height=0.9\textwidth]{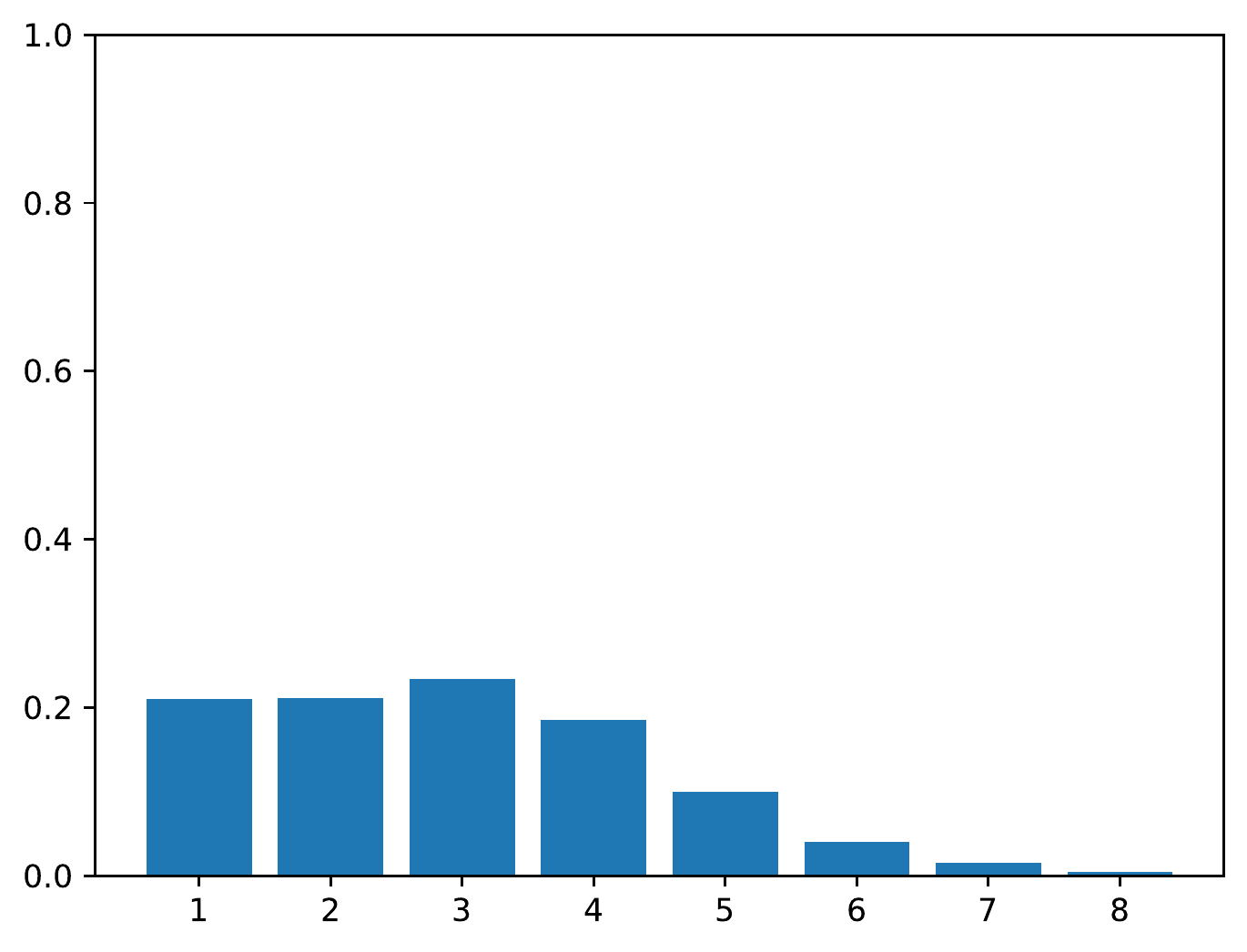}
    \subcaption{SSv2}
    \end{minipage}  
    \vspace{-0.3em}
    \caption{The action start time distribution on three datasets.}
    
    \label{fig:ADM}
\end{figure}
\begin{figure}
    \centering
    \begin{minipage}{0.32\linewidth}
    \includegraphics[width=1\textwidth,height=1.0\textwidth]{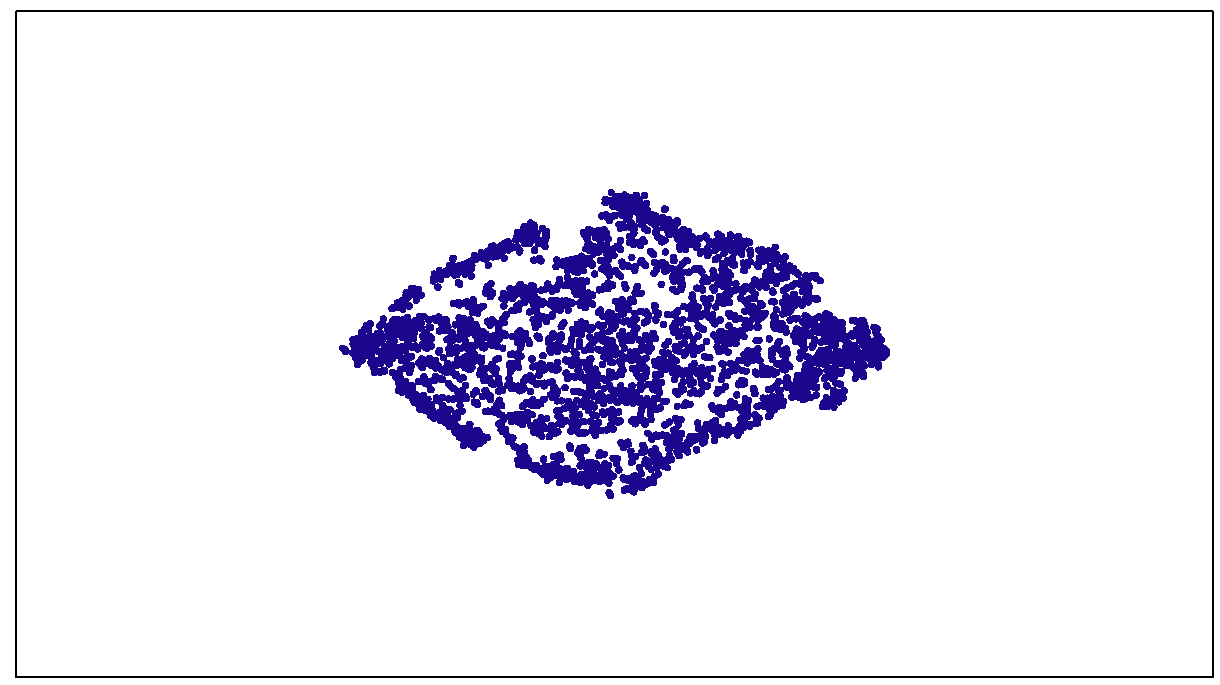}
    \subcaption{UCF101}
    \end{minipage}
    \begin{minipage}{0.32\linewidth}
    \includegraphics[width=1\textwidth, height=1.0\textwidth]{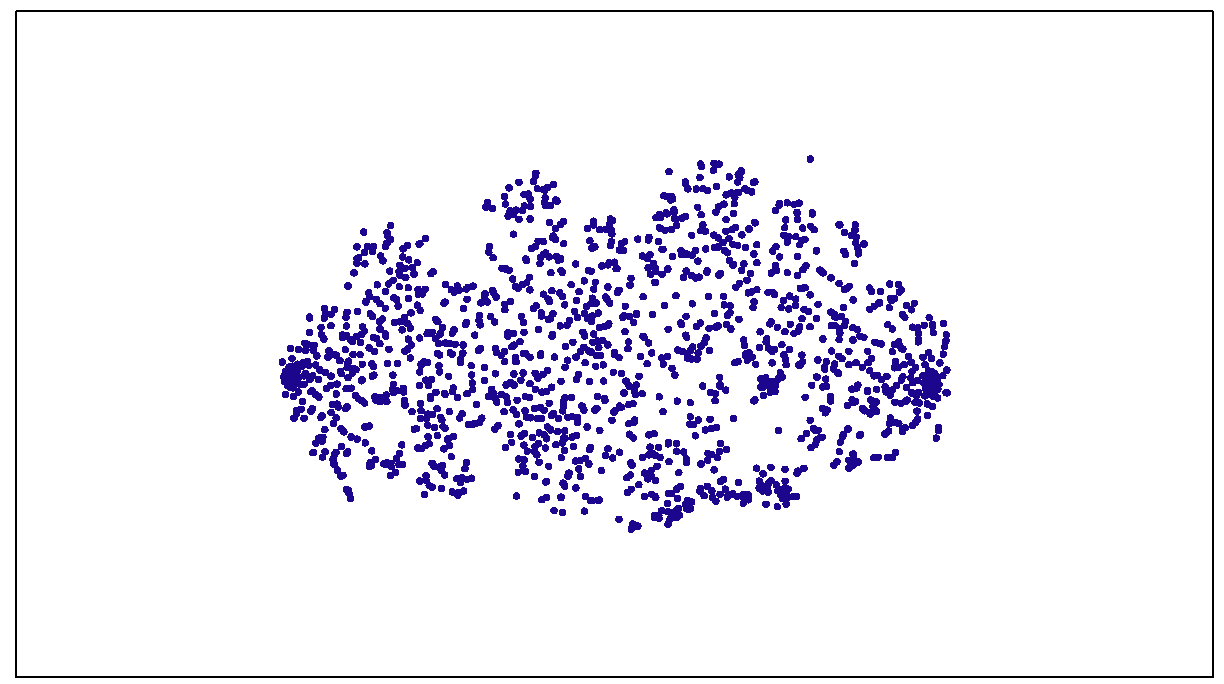}
    \subcaption{HMDB51}
    \end{minipage}
    \begin{minipage}{0.32\linewidth}
    \includegraphics[width=1\textwidth,height=1.0\textwidth]{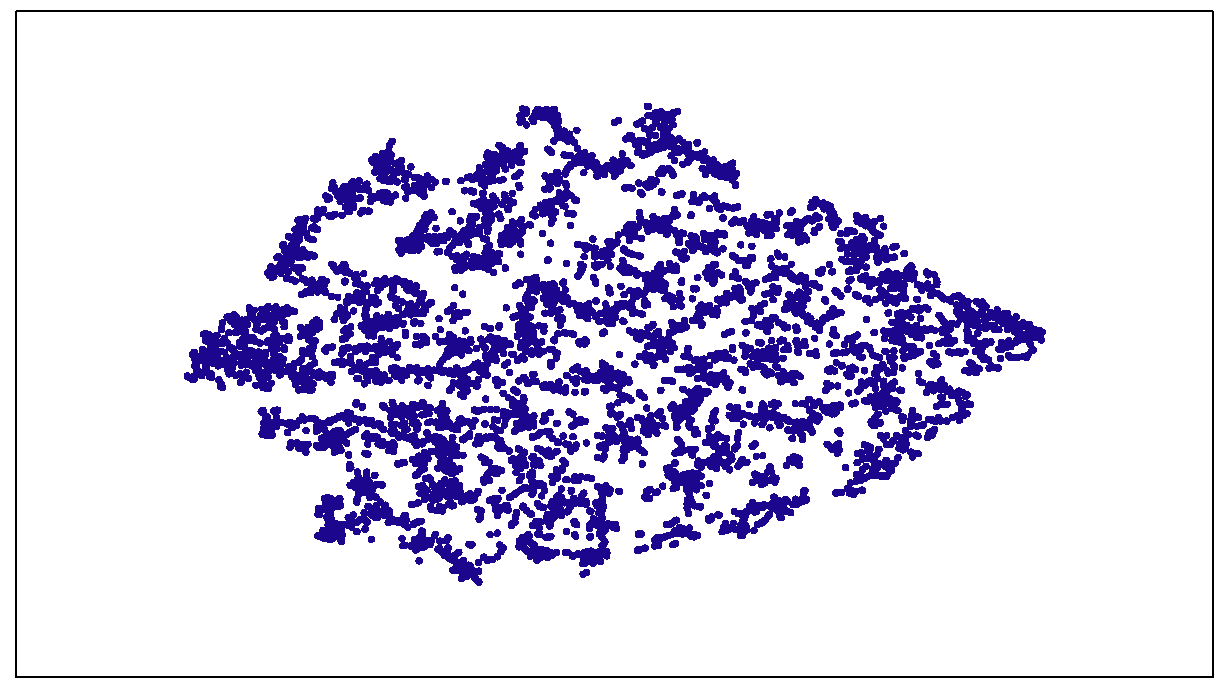}
    \subcaption{SSv2}
    \end{minipage}    
    \caption{2-D t-SNE of the videos on three datasets.}
    \vspace{-0.7em}
    \label{fig:AEM}
\end{figure}
To further analyze the action misalignment, we quantify and compare these two types of misalignment on three video action datasets: (UCF101~\cite{ucf101}, HMDB51~\cite{HMDB}, SSv2~\cite{ssv2}). Please refer to our supplementary for more details about the quantitative method and process. The quantitative results and analysis are discussed as follows.

For the \textit{action duration misalignment (ADM)}, we aim to analyze the distribution of action start time in different datasets, which is presented in Fig.~\ref{fig:ADM}.
For UCF101 and HMDB, the start time is distributed mainly over the first or second frame due to the fact that most videos are roughly trimmed. On the contrary, the start time on the SSv2 dataset is evenly distributed over the first four frames. This demonstrates that the actions on the SSv2 dataset are more likely to execute at various time periods, which could lead to misalignment in the action start and duration time.

For the \textit{action evolution misalignment (AEM)}, we estimate a final AEM score for each dataset by calculating the similarity of action evolution among videos (refer to our supplementary for details about the estimation).
The estimated AEM scores for the three datasets are listed in Tab.~\ref{tab:estimated_mem}. It can be seen that all datasets suffer from the AEM problem. Similar to the ADM, the problem of AEM is the most serious on the SSv2. Among these datasets, the UCF101 has a lower severity of evolution misalignment, since it mainly consists of various types of sports, which provided more consistent action evolution owing to class homogeneity. Furthermore, we visualize the feature embedding of action evolution for all videos to illustrate the degree of evolution misalignment (refer to supplementary for details).
It is apparent that a concentrated distribution is seen on UCF101 and HMDB51 while the distribution is slightly more scattered on the SSv2 dataset. This verifies the similar conclusion that the SSv2 faces a more serious AEM problem.
\begin{table}[t]
\centering
\begin{tabular}{c|c|c|c}
\hline
Dataset                                                 & UCF101 & HMDB51 & SSv2   \\ \hline
\begin{tabular}[c]{@{}c@{}}Estimated\\ AEM\end{tabular} & 0.1653 & 0.3697 & 0.6260 \\ \hline
\end{tabular}
\caption{The estimated action evolution misalignment (AEM) score on three datasets}
\vspace{-0.5em}
\label{tab:estimated_mem}
\end{table}

Overall, from this analysis, we can conclude that the action misalignment problem widely exists in these three datasets at varying levels. The problem is most severe on the SSv2 dataset while HMDB is more affected by this issue than the UCF101. Hence, we argue that solving the action misalignment problem is critical for few-shot action recognition, especially on the SSv2 dataset. Based on these observations, this paper seeks to address the misalignment problem by proposing a feasible framework.

\begin{figure*}[tp]
    \centering
    \subcaptionbox{}{
        \includegraphics[width=0.6\linewidth]{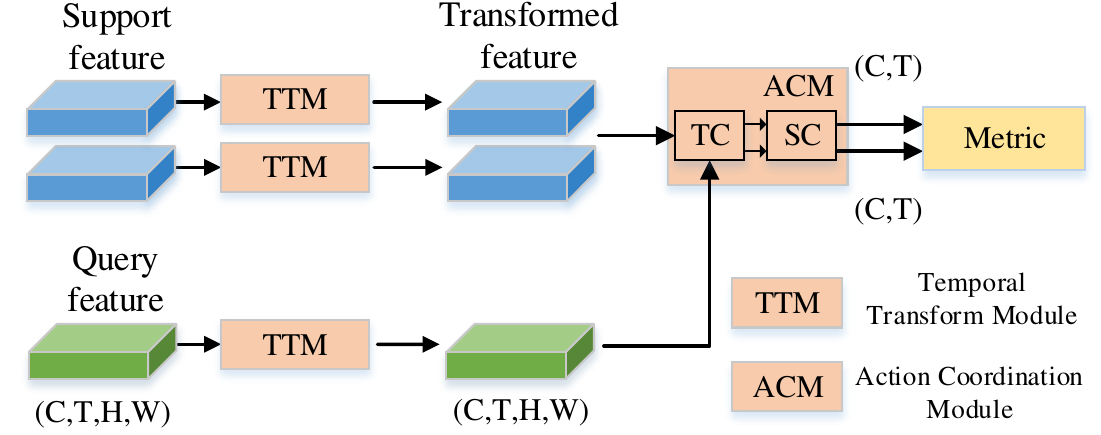}
    }
    \hfill
    {\vrule width0.6pt}
    \subcaptionbox{}{
        \raisebox{6ex}{
            \includegraphics[width=0.35\linewidth]{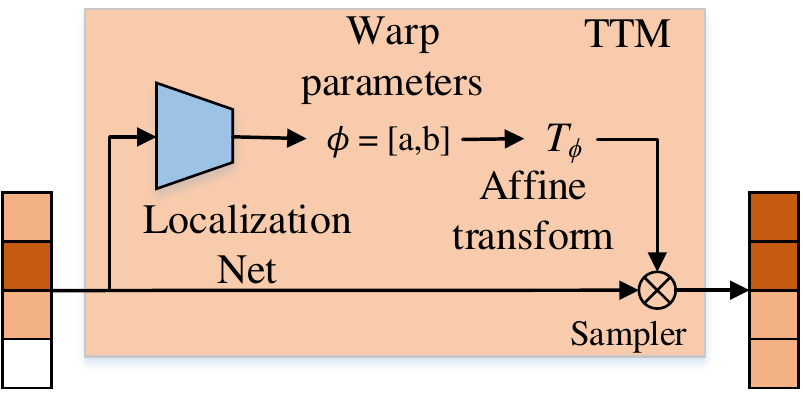}
        }
    }
    \vspace{-0.5em}
    \caption{(a) The proposed framework. We only show 1-way for illustration. Embedded video features are first transformed by TTM (temporal transformation module) to address the action duration misalignment. Then support and query features are finely coordinated by ACM (action coordinate module) along the temporal and spatial dimensions. (b) Structure of TTM.}
    \label{fig:framework}
    \vspace{-1em}
\end{figure*}

\section{Methods}

Fig.~\ref{fig:framework}(a) shows the framework of our method. In the following sections, we formally describe the few-shot action recognition problem, followed by detailed descriptions of modules in our framework. Finally, we describe how to optimize our model.

\subsection{Definition}

\subsubsection{Problem set-up} Following standard few-shot action recognition setting, the dataset is divided into three distinct parts: training set $\mathcal{C}_{train}$, validation set $\mathcal{C}_{val}$, and test set $\mathcal{C}_{test}$. The training set contains sufficient labeled data for each class while there exist only a few labeled samples in the test set. The validation set is only used to evaluate the model during training. Moreover, there are no overlapping categories between these three sets. Generally, few-shot action recognition aims to train a classification network that can generalize well to novel classes in the test set. In the specific $N$-way $K$-shot few-shot learning setting, each episode contains a support set $\mathcal{S}$ sampled from the training set $\mathcal{C}_{train}$. It contains $N \times K$ samples from $N$ different classes where each class contains $K$ support samples. Then Q samples from each class are selected to form the query
set $\mathcal{Q}$ which contains $N \times Q$ samples. The goal is to classify the $N \times Q$ query samples only with the $N \times K$ support samples. 

\subsubsection{Feature embedding}
For each input video, we follow the sampling strategy described in TSN~\cite{TSN2016}, which divides a video into $T$ segments and then samples frames uniformly in each segment. Thus, each video is represented by a fixed-length frame sequence. Given the frame sequence $X=\{x_1, x_2, ..., x_T\}$, a feature embedding network $f(\cdot)$ takes it as input and embeds the sequence $X$ into $T$ frame-level features  $f_X=f(X) = \{f(x_1), f(x_2),\dots,f(x_T)\}\in \mathbb{R}^{\rm C \times T\times H\times W}$. From this point, we will use $f_s$, $f_q$ to represent the video-level feature of the support sample and query sample, respectively.

\subsection{Temporal Transform Module}
\label{sec:TTM}
To address the action duration misalignment, we aim to locate the action temporally, then the duration feature could be located and emphasized while dismissing the action-irrelevant feature (e.g. background). In this way, the ADM could be eliminated. Based on this motivation, we design a Temporal Transform Module (TTM).
It consists of two parts: a localization network $\mathbf{L}$ and an temporal affine transformation $\mathbf{T}$. 

Specifically, given an input frame-level feature sequence $f_X$, the localization network generates warping parameters $\phi=(a,b)=\mathbf{L}(f_x)$ firstly. Then the input feature sequence is warped by the affine transformation $\mathbf{T}_{\phi}$. 
Succinctly, the temporal transform process is defined as:
\begin{equation}
    \hat{f_X} = \mathbf{T}_{\phi}(f_X),~ \phi=\mathbf{L}(f_X)
\end{equation}
where $\hat{f_X}$ indicates the feature sequence aligned to the action duration period, $\mathbf{L}$ consists of several light trainable layers in our implementation. Since the action duration misalignment is characteristically \textit{linear} among frame sequences, the warping is represented using linear temporal interpolation. This also facilitates the entire pipeline differentiable and thus we can jointly train our classifier with TTM in an end-to-end manner. 

The framework of TTM is illustrated in Fig.~\ref{fig:framework}(b). 
During the episode training and testing, all the feature sequences of the support and query samples are first fed into TTM to perform first-stage temporal alignment, where their video features could be roughly aligned to their action periods. In this way, the TTM stage caters towards relieving the action duration misalignment problem.

\subsection*{Action Coordinate Module}
The second type of misalignment, action evolution misalignment, results from the non-linear evolution of actions in videos, which cannot be adequately addressed by the linear-based TTM. 
To this end, we coordinate action evolution among videos from temporal and spatial aspects.

\begin{figure}[t]
    \centering
    \includegraphics[width=0.95\linewidth]{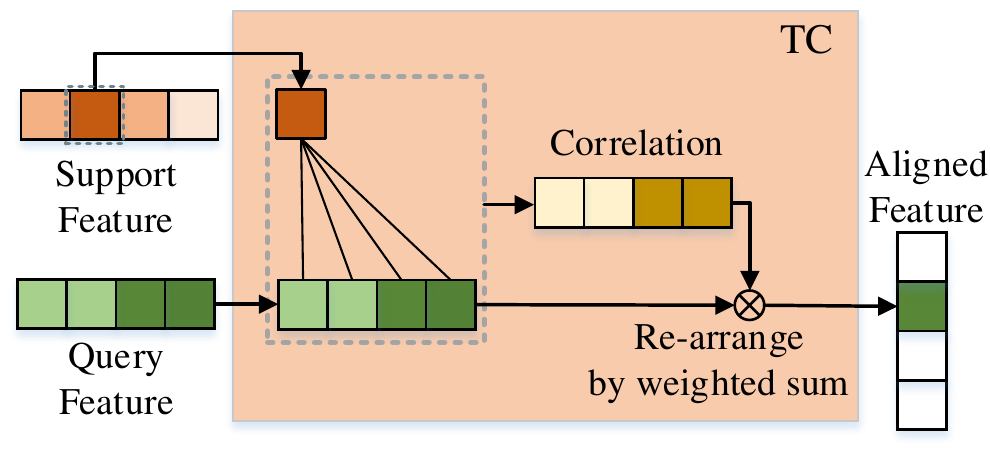}
    \vspace{-0.5em}
    \caption{Temporal Coordination (TC)}
    \label{fig:TC}
    \vspace{-0.9em}
\end{figure}

\subsubsection{Temporal coordination} To temporally align the action evolution among videos, similar motion patterns between videos should be aggregated to the same temporal location. We treat this as a global coordination task where the motion evolution of query video can be temporally rearranged to match the support ones. Accordingly, we model the motion evolution correlation $M\in \mathbb{R}^{T\times T}$ between 
support and query:
\vspace{-0.3em}
\begin{equation}
 M = \operatorname{Softmax}(\frac{(W_k\cdot G(\hat{f_s}))(W_q\cdot G(\hat{f_q}))^T}{\sqrt{dim}})
 \label{eq:tam}
\end{equation}
where $W_k, W_q$ indicate linear projection layer, $dim$ is the dimension of feature $G(\hat{f})$, $G$ is the global average pooling in spatial dimensions whose output tensor shape is $C\times T\times 1 \times 1$, \emph{i.e.} the correlation is only calculated in the temporal dimension, $\operatorname{Softmax}$ limits the values in $M$ to $[0,1]$.

Then, we could temporally rearrange the query feature by calculating the matrix multiplication between the normalized motion correlation matrix $M$ and the query features:
\begin{gather}
\tilde{f_q}= M\cdot (W_v\cdot G(\hat{f_q}))
\end{gather}
Similar to $W_k,W_q$, $W_v$ denotes linear projection layer. In order to keep feature-space consistent, this projection are also applied to support feature $\hat{f_s}$: $\tilde{f_s}= W_v\cdot G(\hat{f_s})$. This way, the same temporal location is in the consistent evolution and AEM in temporal aspect can be relieved. The illustration of TC is presented in Fig.~\ref{fig:TC}.

\subsubsection{Spatial coordination}
\begin{figure}
    \centering
    \includegraphics[width=0.9\linewidth]{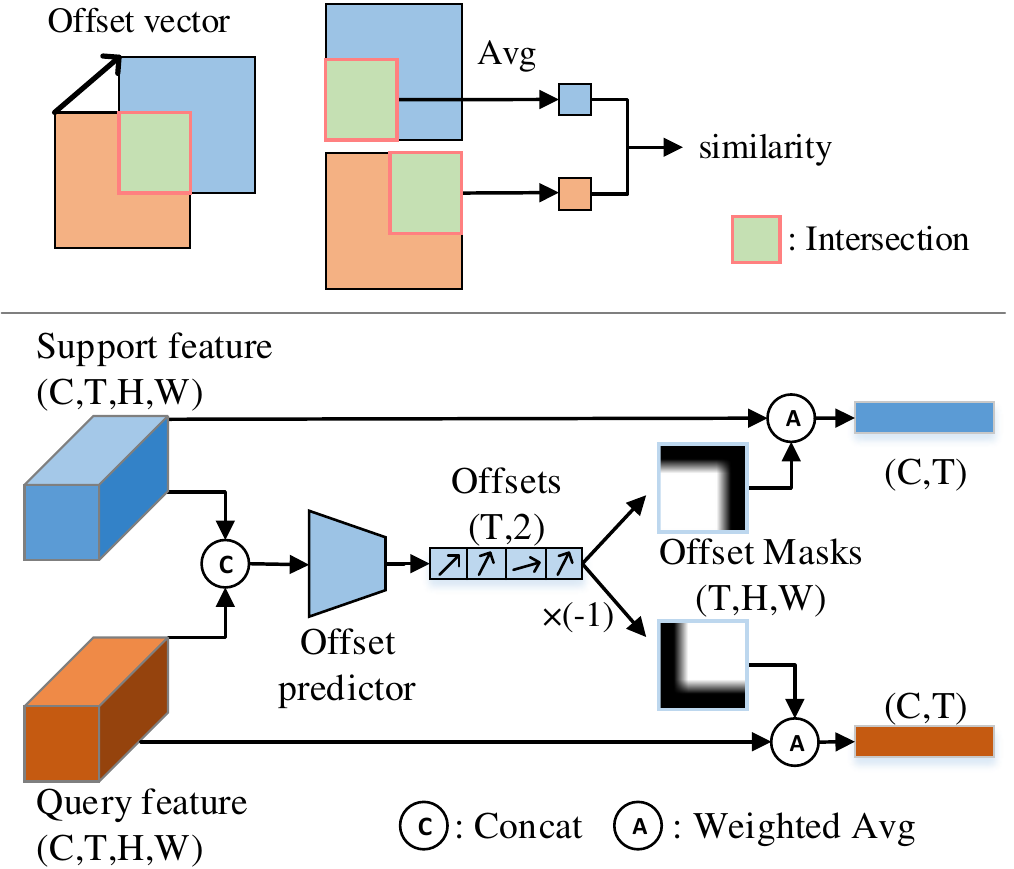}
    \caption{Spatial coordination (SC). \textit{top}: illustration of how to measure the distance under the offset. \textit{bottom}: details of offset prediction and mask generation.}
    \label{fig:SC}
    \vspace{-1.1em}
\end{figure}

Temporal coordination (TC) ensures the action evolves with same process along the duration time. However, the spatial variation of actor evolution, such as the positions of actors, also being critical for action recognition, which cannot be modeled by TC. Thus, we further devise a spatial manipulation to reduce the spatial variations in action evolution. On the basis of temporally well-aligned features, we aims to predict an spatial offset for each paired frames and then measure their similarity in the intersection area only, as shown in the top of Fig.~\ref{fig:SC}.

Specifically, spatial coordination consists of two steps: light-weight offset prediction and offset mask generation.
First, given the temporally well-aligned feature $\tilde{f_s}$ and $\tilde{f_q}$, they are feed into the offset predictor $S$ to predict spatial offset $O\in \mathbb{R}^{T\times 2}$ in $x$ and $y$ coordinates for all timestamps:
\begin{gather}
O=S(\operatorname{Cat}(\tilde{f_s}, \tilde{f_q}))
\end{gather}
where $\operatorname{Cat}(\cdot)$ denotes concatenation along the channel. The detailed structure of $S$ is elaborated in our Supplementary Material. As shown in Fig.\ref{fig:SC}, the predicted offset indicates the relative position vector of the action-specific region between query and support frames. Then, the intersection area can be located by its corresponding spatial offset.

To calculate the similarity in the intersection area in a differentiable way, for each frame, we use a generated offset mask $I$ to calculate the average feature in the intersection on each feature. Moreover, the value of the mask is 1 inside the intersection area and it gradually decreases to 0 on the edge.
More details about mask generation are provided in the supplementary material.
Then the masks are performed over the query and support features simultaneously as the weights of weighted average:
\begin{gather}
\overline{f}_{s,i}=\sum_{HW}{(I_{o_i}*\tilde{f_s})} / \sum{I_{o_i}},~i=0,\dots,T\\
\overline{f}_{q,i}=\sum_{HW}{(I_{-o_i}*\tilde{f_q})} / \sum{I_{-o_i}},~i=0,\dots,T
\end{gather}
where $I_{o_i}$ is the generated mask for $i$-th frame. 

In order to expand the exploration space of the offset predictor, we further add some perturbations on predicted offset and use the average of corresponding features of different perturbations.

Undergoing TC and SC, the action evolution misalignment among videos are dismissed in spatial-temporal aspect.
The well-aligned paired features $\overline{f_s}$ and $\overline{f_q}$ are used in final distance measurement and classification as the prototypical network scheme~\cite{prototypical2017}.



\subsection{Optimization}

We train our model in similar manner as the ProtoNet~\cite{prototypical2017} framework with standard softmax cross-entropy. 
Given the aligned feature of query sample $\overline{f}_q$, and the support prototype $\overline{p}_s^c$ 
for class $c$ (obtained by applying TC to k-shot support features and average, refer to supplementary for details), we can obtain the classification probability as:
\begin{equation}
    {\rm P}(x_q \in c_i)=\frac{\exp(-d(\overline{f_q},\overline{p}_s^{c_i}))}{\sum_{c_j \in C}{\exp(-d(\overline{f_q},\overline{p}_s^{c_j}))}}
\end{equation}
\begin{equation}
    d(f,p)=\sum_{t=1}^{T}{1- \frac{<f_{[t]},p_{[t]}>}{\|f_{[t]}\|_2 \|p_{[t]}\|_2} },
\end{equation}
where $d(f,p)$ is the frame-wise cosine distance metric.
Then the classification loss is calculated as:
\begin{equation}
    \mathcal{L}_{cls}= -\sum_{q\in Q}{\mathbb{I}(q\in c_i)\log{{\rm P}(x_q \in c_i)}},
\end{equation}
where $\mathbb{I}$ is an indicator function. $d$ denotes the distance metric whereby we adopt the time-wise cosine distance in our implementation. $Q$ and $C$ represent the query set and its corresponding collection of class label, respectively.

\begin{table*}[t]
\centering
\begin{tabular}{c|c|cc|cc|cc|cc}
\hline
\multirow{2}{*}{Method} &
\multirow{2}{*}{Backbone} &
\multicolumn{2}{c|}{HMDB51} & \multicolumn{2}{c|}{UCF101} & \multicolumn{2}{c|}{SSv2} & \multicolumn{2}{c}{Kinetics-CMN} \\
 & & 1-shot & 5-shot & 1-shot & 5-shot & 1-shot & 5-shot & 1-shot & 5-shot \\ \hline
CMN & ResNet-50 &- & - & - & - & - & - & 60.5 & 78.9 \\
TARN & C3D& - & - & - & - & - & - & 64.8 & 78.5 \\
ARN & 3D-\textit{464}-Conv& 45.5 & 60.6 & 66.3 & 83.1 & - & - & 63.7 & 82.4 \\
ProtoNet & ResNet-50& 54.2 & 68.4 & 74.0 & 89.6 & 33.6 & 43.0 & 64.5 & 77.9 \\
TRN++ & ResNet-50& - & - & - & - & 38.6 & 48.9 & 68.4 & 82.0 \\
OTAM* & ResNet-50& 54.5 & 66.1 & 79.9 & 88.9 & 42.8 & 52.3 & 73.0 & 85.8 \\ \hline
Ours & ResNet-50& \textbf{59.7} & \textbf{73.9} & \textbf{81.9} & \textbf{95.1} & \textbf{47.6} & \textbf{61.0} & 72.8 & \textbf{85.8} \\ \hline
\end{tabular}
\caption{Few-shot action recognition results under standard 5-way k-shot settings. Note: * means our implementation.}
\label{tab:compare}
\vspace{-0.75em}
\end{table*}

\section{Experiments}
\subsection{Datasets and Baselines}

\noindent
We conduct experiments on four benchmark datasets:
\begin{itemize}
    \item \textit{UCF101}~\cite{ucf101}:
    We follow the same protocol introduced in ARN~\cite{ARN2020}, where 70/10/21 classes and 9154/1421/2745 videos are included for train/val/test respectively.
    \item \textit{HMDB51}~\cite{HMDB}: Each category contains at least 101 videos. We also follow the protocol of ARN~\cite{ARN2020}, which takes 31/10/10 action classes with 4280/1194/1292 videos for train/val/test.
    \item \textit{SSv2}~\cite{ssv2}: We adopt the same protocol as~\cite{otam2020} where 64/12/24 classes and 77500/1925/ 2854 videos are included for train/val/test respectively.
    \item \textit{Kinetics-CMN}~\cite{compound2018} contains 100 classes selected from Kinetics-400, where 64/12/24 classes are split into train/val/test set with 100 videos for each class.
\end{itemize}

\noindent
\textbf{Competing methods} We compare our method against state-of-the-art FSL action recognition methods related to temporal handling, including ProtoNet~\cite{prototypical2017}, CMN-J~\cite{cmn-j2020}, TARN~\cite{tarn2019}, ARN~\cite{ARN2020}, TRN++, and  OTAM~\cite{otam2020}.

\subsection{Implementation Details}
To be specific, 5-way 1-shot and 5-way 5-shot classification tasks are conducted on all datasets. For all datasets, we sample 8 frames uniformly for each video in the standard way introduced by TSN~\cite{TSN2016}. Extracted frames are first resized to 256$\times$256 and random horizontal flip is applied. 
Then random crop with size 224$\times$224 is applied during training.
We use the ImageNet pre-trained ResNet-50~\cite{he2016deep} as the feature extractor so that we could have a fair comparison with previous methods~\cite{otam2020}. Specifically, the feature before the last average pooling layer in ResNet-50 forms the frame-level input to our TA$^2$N. During meta-training, we sample 200 episodes in a single epoch.  In testing phase, we sample 5000 episodes in the meta-test spilt and report the average result. For more details about training strategies (\emph{e.g.} optimizer, learning rate), please refer to supplementary material.

\subsection{Main Results}

\subsubsection{\textbf{Quantitative Results}}
The quantitative results are listed in Tab.~\ref{tab:compare}. 
As shown in this table, our method outperforms the strong baseline ProtoNet~\cite{prototypical2017} on all datasets and is competitive with state-of-the-art methods. OTAM is the current state-of-the-art method that focuses on temporal alignment (and the most related to ours). Compared to it, TA$^2$N surpasses its performance by a significant margin on most settings and datasets, demonstrating the superiority of our proposed two-stage alignment framework for action alignment. Moreover, our TA$^2$N aligns actions in the spatial-temporal aspect while OTAM only considers the temporal dimension. 

Among four benchmarks, the TA$^2$N gains the most significant improvement on the SSv2. This finding also concurs with our quantitative analysis on the state of misalignment in datasets, whereby SSv2 manifests the most serious misalignment problem. This further demonstrates the effectiveness of our alignment modules. Although the UCF101 possesses a relatively less serious condition, our  TA$^2$N could still improve its performance by learning a more consistent temporal feature.

\subsubsection{\textbf{Qualitative Results and Visualizations}}
\begin{figure*}[t]
    \centering
    \includegraphics[width=0.9\linewidth]{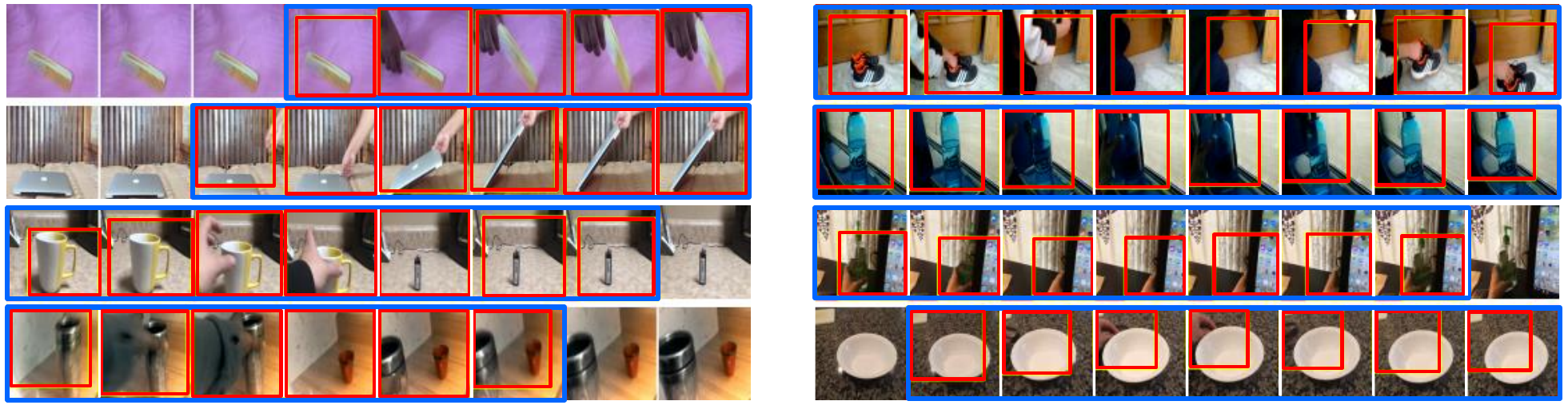}
    \caption{The visualization results on SSv2. Action duration (highlight with blue boxes) is obtained by predicted warp parameters in TTM. Action-specific spatial region (highlight with red box) is located with predicted offset in SC. All presented video clips are taken from the test set.
    }
    \label{fig:vis_align}
    \vspace{-0.75em}
\end{figure*}

We visualize the alignment results to illustrate the effectiveness of our proposed method, which are presented in Fig.~\ref{fig:vis_align}. It can be observed that there exists a clear duration and spatial-temporal evolution misalignment between the support and query videos. It's clear that the duration is well-aligned by the TTM, which filters the insignificant background frame noise. Besides, the spatial regions coordinated by SC focus on the common action-specific part between paired frames. For example, the region of hand (row 3 col 2 in Fig.~\ref{fig:vis_align}) and the action-specific object `cup' (row 3 col 1 in Fig.~\ref{fig:vis_align}) can be located precisely, which leads to a well-aligned spatial evolution for videos before being compared. 

In summary, the visualization tellingly depicts the capability of TA$^2$N in correcting the misalignment.

\subsection{Ablation Study}

\begin{table}[t]
\small
\centering
\begin{tabular}{ccc|cc}
\hline
\multirow{2}{*}{TTM} & \multicolumn{2}{c|}{ACM} & \multirow{2}{*}{UCF101} & \multirow{2}{*}{SSv2} \\
           & TC        & SC        &      &      \\ \hline
           &            &            & 74.0 & 40.1 \\
\checkmark &            &            & 78.5 & 43.8 \\
\checkmark & \checkmark &            & 80.9 & 46.3 \\
\checkmark &  &     \checkmark       & 79.8 & 45.3 \\
\checkmark & \checkmark & \checkmark & \textbf{81.9} & \textbf{47.6} \\
           & \checkmark &            & 78.5 & 44.8 \\
            & \checkmark & \checkmark & 81.0 & 47.0 \\          
 \hline
\end{tabular}
\caption{Ablation of different modules, reported under 5-way 1-shot setting. \checkmark means module is applied.}
\label{tab:breakdown}
\vspace{-1em}
\end{table}

\subsubsection{\textbf{Breakdown Analysis}}
Firstly, we break down our proposed TA$^2$N into its component parts and compare the performance gain of the TTM and ACM when applied separately. Quantitative results on UCF101 and SSv2 datasets are listed in Tab.~\ref{tab:breakdown}. We can observe that both TTM and ACM can boost the performance of few-shot action recognition with each stage playing an equally important role in action alignment. When TTM and ACM are applied in a two-stage manner, the performance is further improved, thereby supporting our notion of a two-stage sequential design.

Further, we split the spatial and temporal coordination parts in ACM. A single TC improves the baseline with 4.5 and 4.7 in UCF and SSv2 respectively. When TTM and TC are applied sequentially, the performance grows with a large margin. This proves that our proposed TTM and TC could well address the \textit{temporal} misalignment from two distinctive aspects. Combined with SC, we can obtain the best performance, which advocates the necessary of aligning the evolution in \textit{spatial} dimension. In a word, the above results point towards the fact that our TA$^2$N provides an effective solution to address these two critical misalignment problems (as handled in the two stages) in few-shot action recognition task.


\subsubsection{Design of spatial coordination} To reveal the effectiveness of spatial coordination (SC) module design, we compare our implementation (\textit{Mask-based}) with other alternative designs. (1) \textit{Enumerate}: it enumerates through all possible integer offsets in $\mathbb{Z}^{2}$, and straightly index the intersection area in feature $x,y$ coordinates.
The offset with the minimum metric distance between support and query is considered the optimal one.
It can be regarded as a simple baseline. (2) \textit{Grid-based}: it generates grids according to our predicted offset and then uses the re-sampling trick to sample features in the intersection area. 
Their results are shown in Tab.~\ref{tab:aba:spatial}. As we can see, our implementation outperforms the simple enumeration by a great margin, which also demonstrates the ability of our offset predictor in SC. Compared to \textit{Grid-based} manner, our design is more straightforward and computationally tractable, with slightly better performance. 

\begin{figure}[t]
    \centering
    \includegraphics[width=0.48\linewidth]{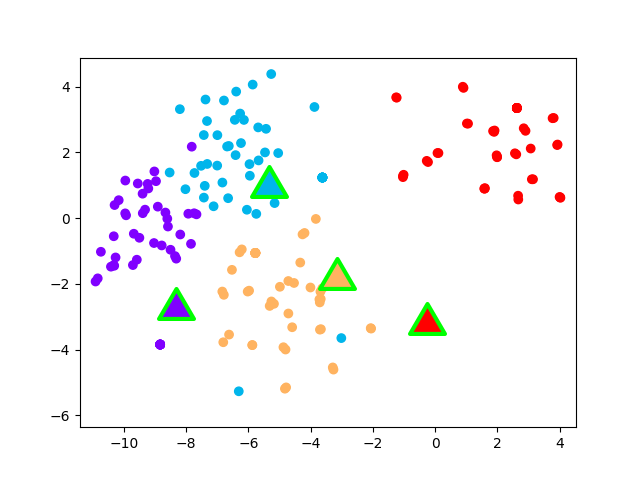}
    \hfill
    \includegraphics[width=0.48\linewidth]{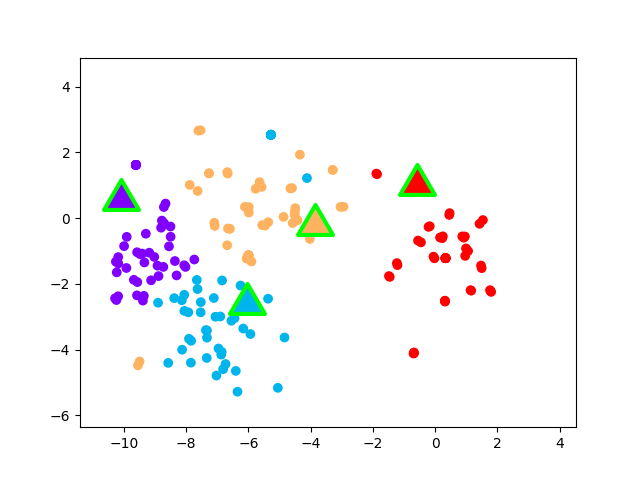}
    \caption{t-SNE feature embedding of query videos (\textit{circles}) and support prototypes (\textit{triangles}) on HMDB. Left/Right: before/after applying our alignment.}
    \label{fig:tsne}
    \vspace{-1.15em}
\end{figure}

\subsubsection{t-SNE visualization} To further illustrate the effect of our proposed TA$^2$N more intuitively, we visualize the feature embedding of query and the prototype of support before and after applying our framework by t-SNE method in Fig.~\ref{fig:tsne}. We can see that each cluster appears more concentrated and closer to its prototype after alignment. It further proves that our TA$^2$N can well align the query videos to support videos and obtain a more consistent feature representation.

\begin{table}[t]
\small
\centering
\begin{tabular}{c|cc}
\hline
 & HMDB51 & UCF101 \\ \hline
Enumerate & 57.6 & 80.8 \\
Grid-based & 59.7 & 81.4 \\
Mask (Ours) & \textbf{59.9} & \textbf{81.9} \\ \hline
\end{tabular}
\caption{Accuracy of different designs of SC, reported under 5-way 1-shot setting.}
\label{tab:aba:spatial}
\vspace{0.5em}
\end{table}

\subsubsection{Class-specific improvement} 
\begin{figure}[t]
    \centering
    \includegraphics[width=\linewidth]{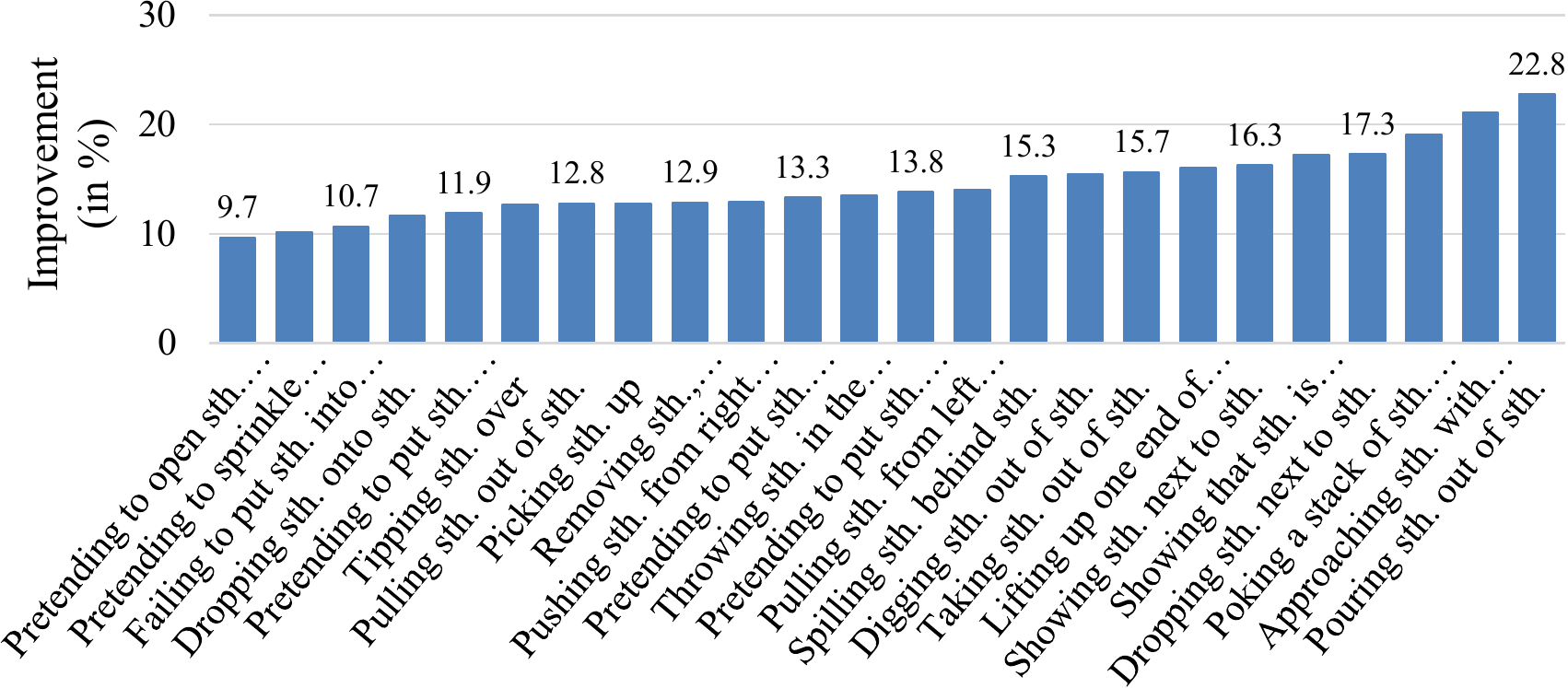}
    \caption{Class-specific improvement using TA$^2$N compared to the Prototype baseline for SSv2. }
    \label{fig:class_improvement}
    \vspace{-1.25em}
\end{figure}

The improvement on the SSv2 dataset for each category using the proposed TA$^2$N is presented in Fig.~\ref{fig:class_improvement}. What stands out in this figure is that the performance increases with a large margin for all categories. Moreover, some categories' accuracy (\emph{e.g.} ``pouring sth out of sth'', ``approaching sth'', ``poking a stack of sth'') rose sharply ($>20\%$ improvement) using TA$^2$N. Interestingly, these action classes are the ones that are more vulnerable to the misalignment problem. It also prove the advantages of 
temporal alignment in actions with limited information.

\section{Conclusion}
This paper delves into the inevitable issue of action misalignment in few-shot action recognition and proposes a new Two-stage Action Alignment Network (TA$^2$N) to address it. Its benefits rests on 
two action alignment modules -- the first performs temporal transformation to handle misalignment in duration, the second performs temporal rearrangement and spatially offset prediction to coordinate the evolution of action between video feature pairs. 
Extensive experiments affirm the effectiveness of the proposed framework. 

\section{Acknowledgements}
This work is supported in part by the following grants: National Key Research and Development Program of China Grant (No.2018AAA0100400), National Natural Science Foundation of China (No. U21B2013, 61971277).

\newpage

\bibliography{aaai22}

\newpage
\maketitle
\section{Supplementary Material}

\subsection{Quantifying Misalignment}
To quantify the action misalignment among three video datasets, we firstly train a general action recognition model based on TSM~\cite{TSM2019} for each dataset separately. Given the well-trained classifier, we fed videos with $T$ uniformly sampled frames into it and obtain the frame-level class probability vector $\mathcal{P}\in \mathcal{R}^{C\times T}$, where $C$ denotes the number of classes. For each video, we extract the class-specific probability sequence $P=[p_1,\dots,p_T] \in \mathcal{R}^{T}$ from $\mathcal{P}$ according to its ground-truth category. This class-specific probability vector represents the evolution of action within $T$ frames. 

\textit{\textbf{Action start time distribution}}: For each video, we consider the frame index $t$ that satisfies $p_{t}\geq 0.5$ and $p_{1,\dots,t-1}<0.5$ as the \textit{start time}. 

\textit{\textbf{AEM estimation}}: Based on the probability sequence, we calculate the cosine similarity for all video pairs in each dataset. Then, we compute action evolution misalignment (AEM) for each dataset by the following operation:
\begin{equation}
    AEM = \frac{1}{M}\sum_{i,j}[1-cos(P_i,P_j)],~\forall (i,j) 
\end{equation}
where $M=2\cdot v_{num} \cdot(v_{num}-1)$ is the normalization coefficient, $v_{num}$ is the number of videos in dataset, $cos(\cdot)$ calculates the cosine similarity for the pair of inputs $(i,j)$. 


\begin{figure*}[t]
    \centering
    \includegraphics[width=0.45\linewidth]{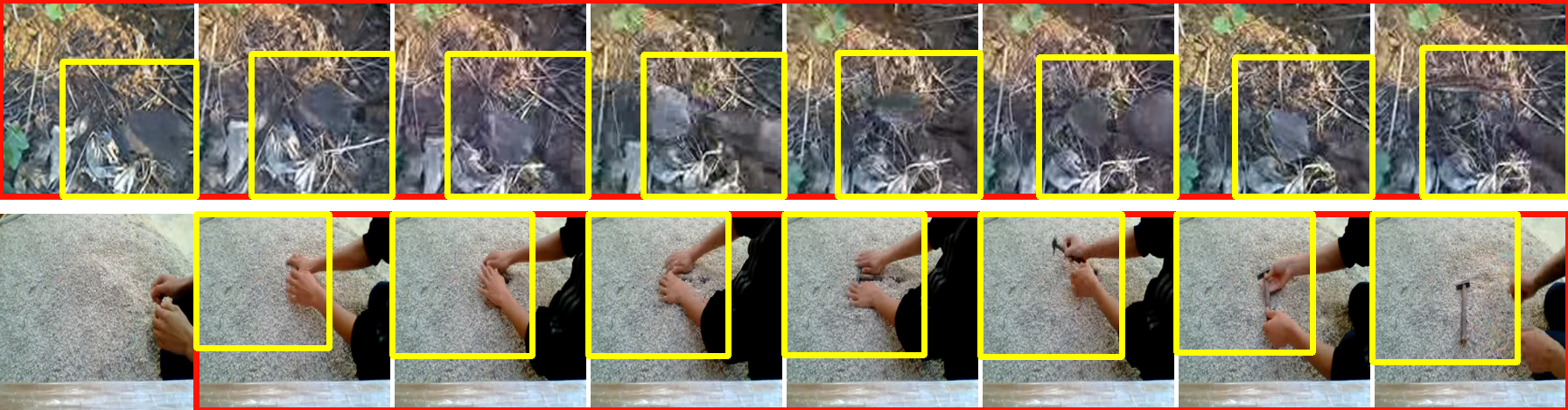}
    \includegraphics[width=0.45\linewidth]{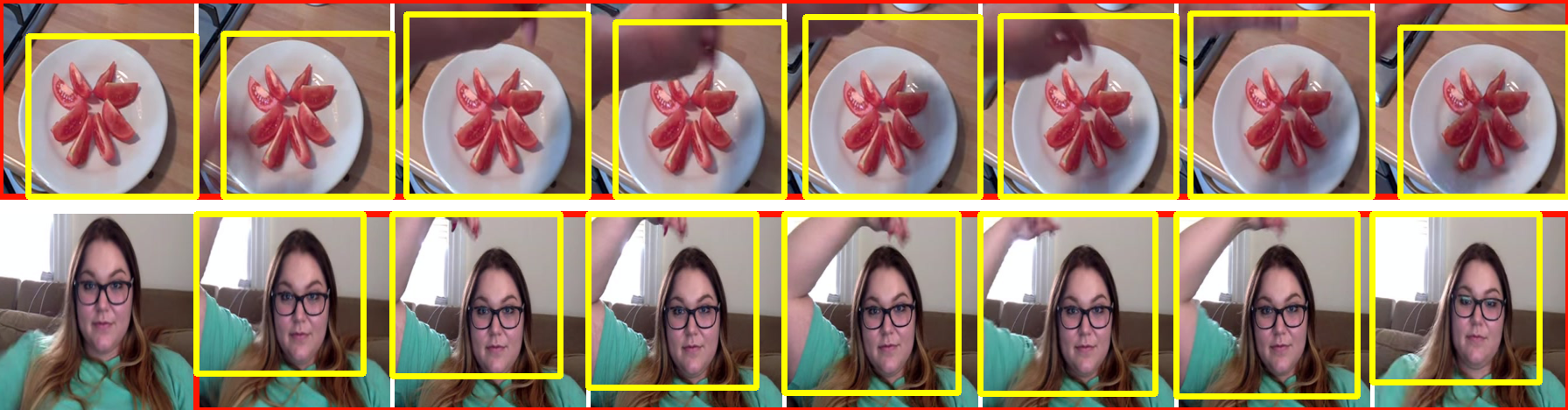}\\
    \includegraphics[width=0.45\linewidth]{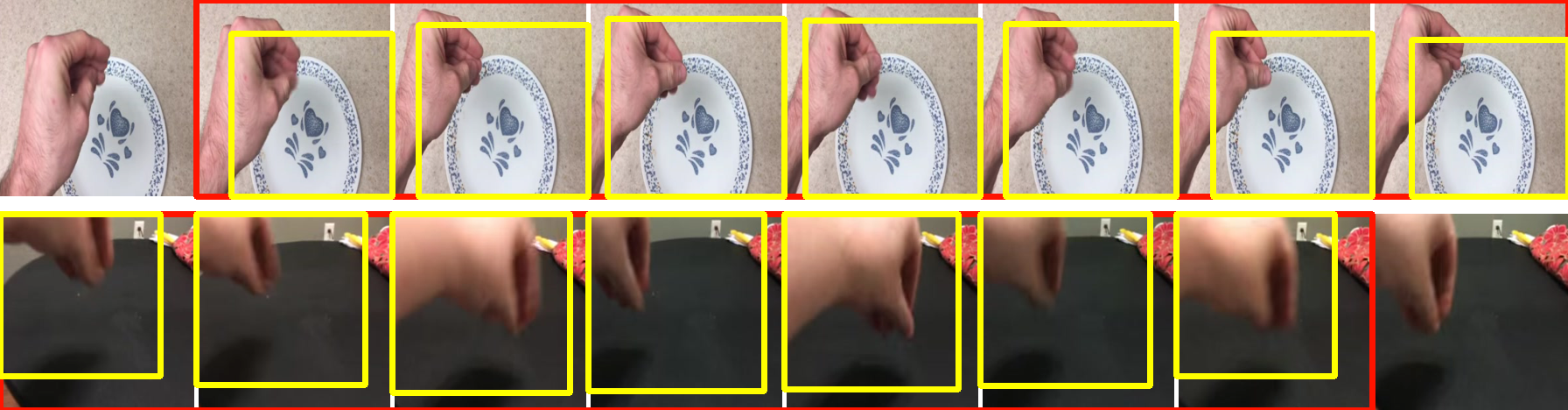}
    \includegraphics[width=0.45\linewidth]{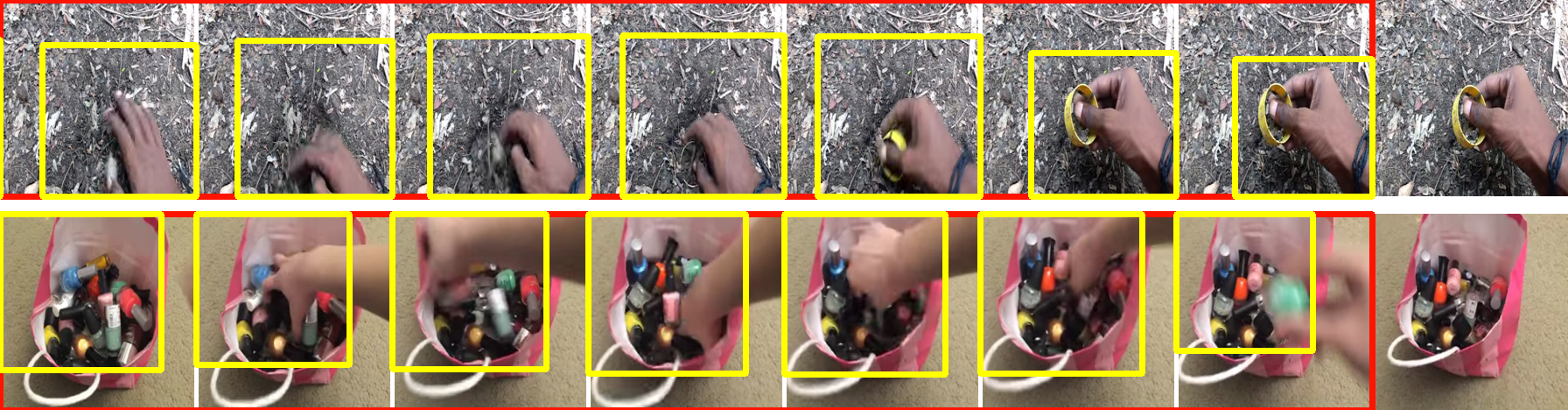}\\
    \includegraphics[width=0.45\linewidth]{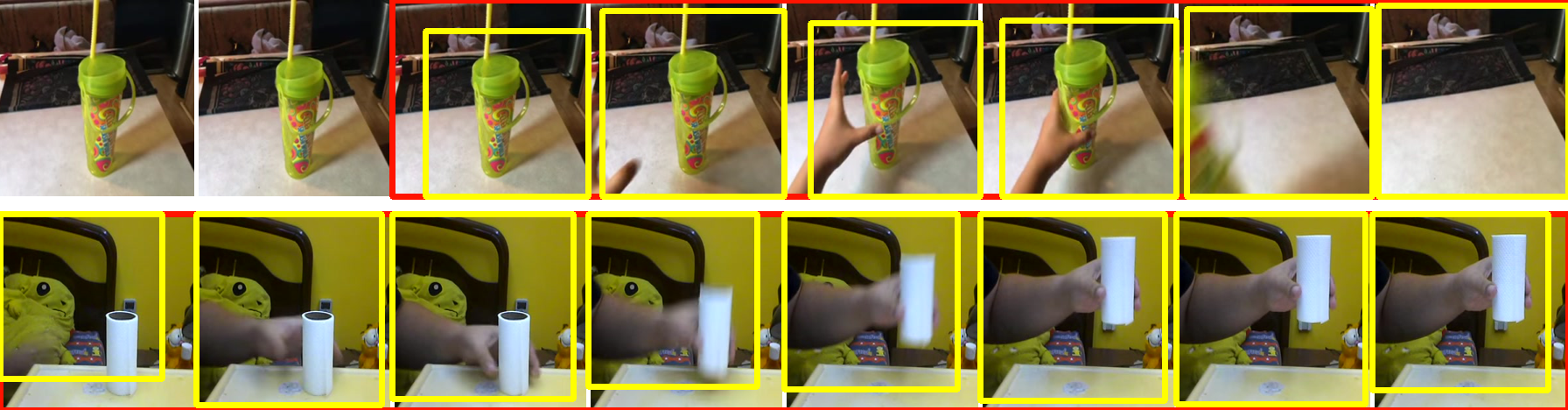}
    \includegraphics[width=0.45\linewidth]{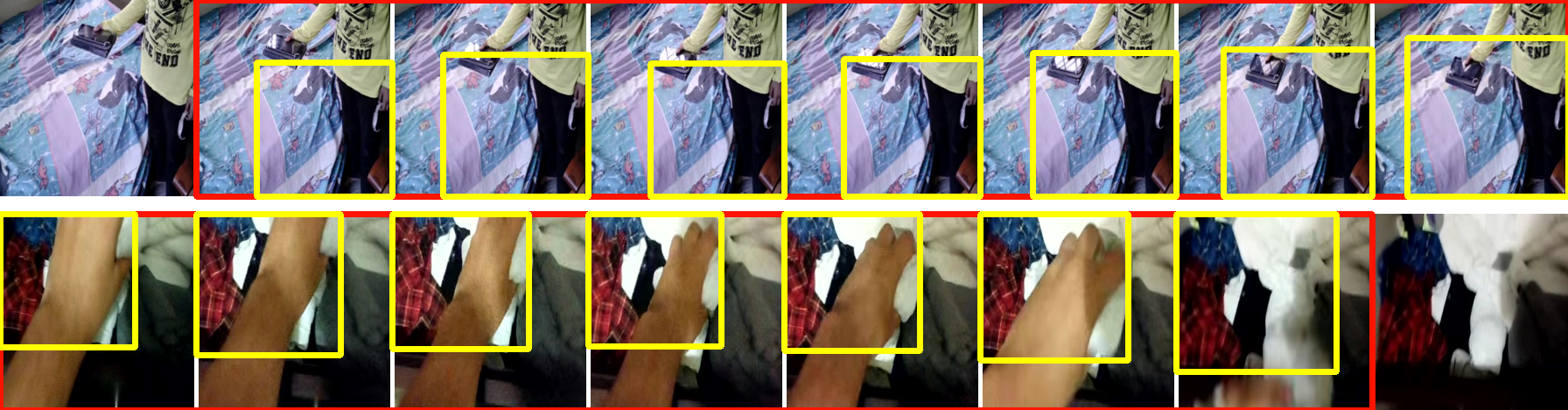}\\
    \includegraphics[width=0.45\linewidth]{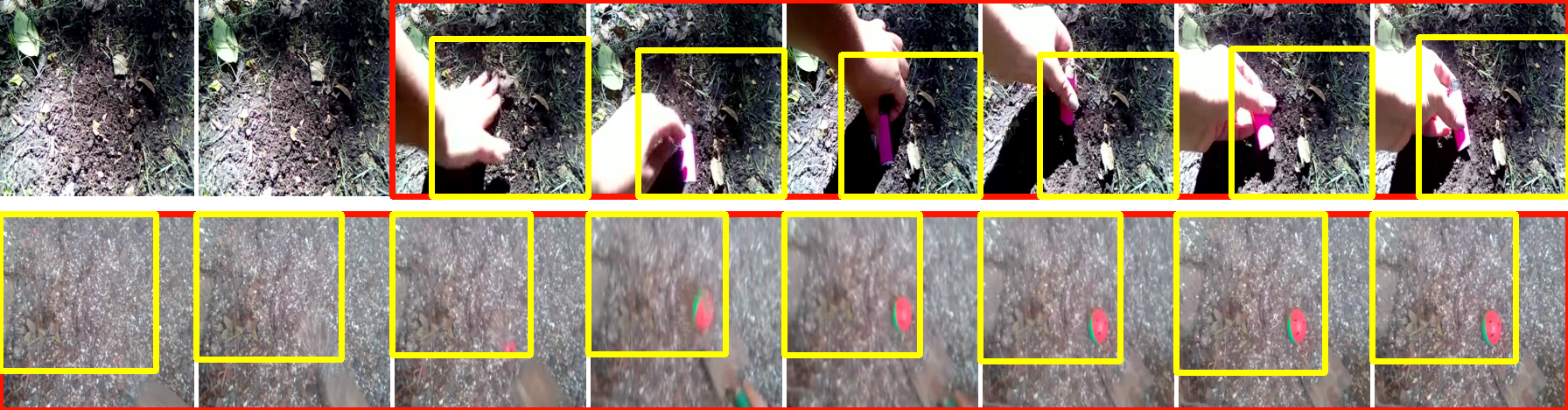}
    \includegraphics[width=0.45\linewidth]{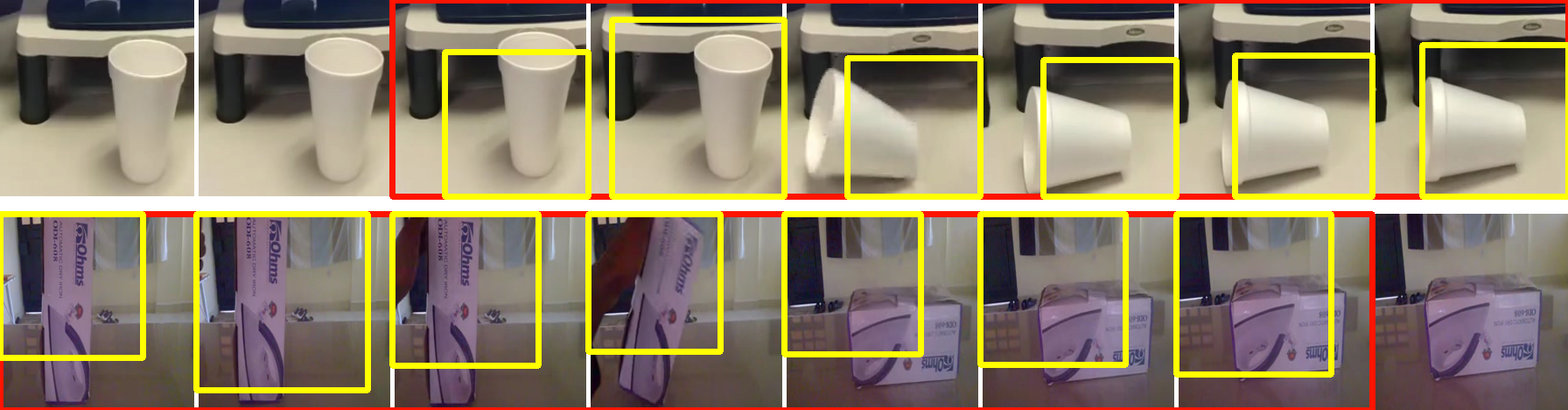}\\
    \includegraphics[width=0.45\linewidth]{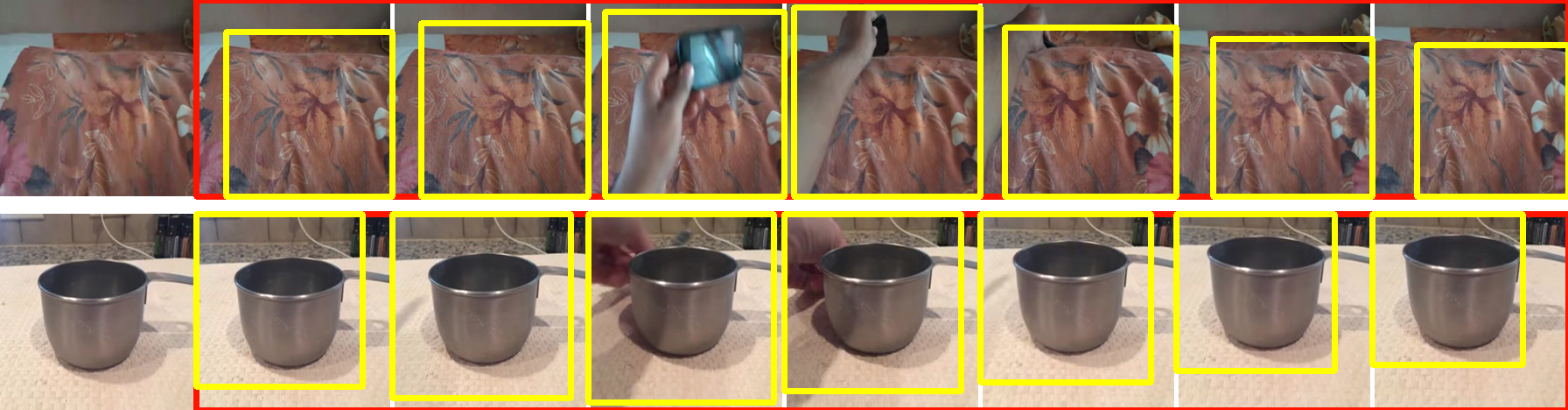}
    \includegraphics[width=0.45\linewidth]{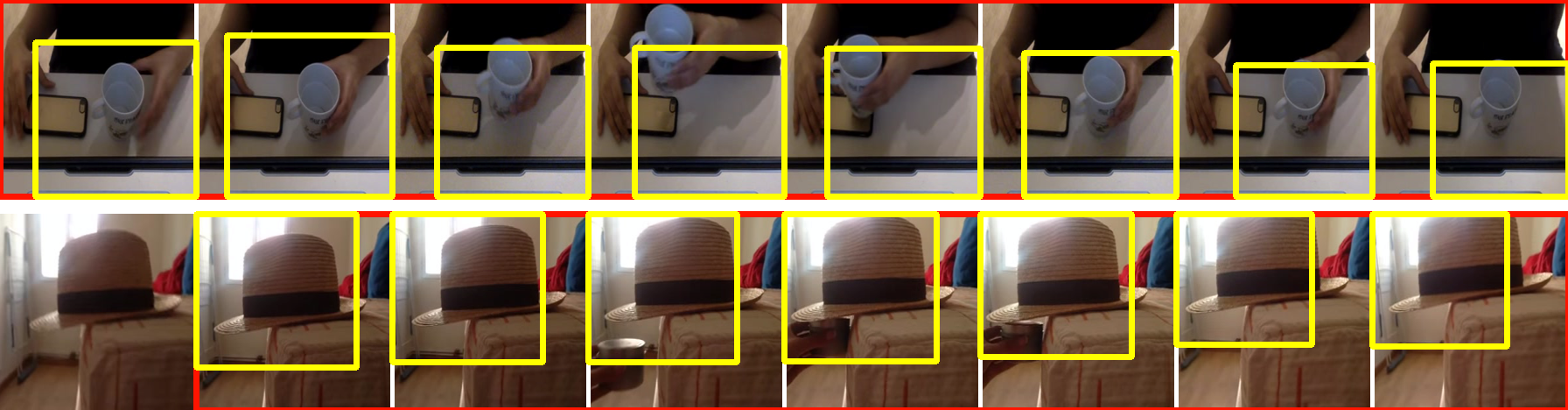}\\
    \includegraphics[width=0.45\linewidth]{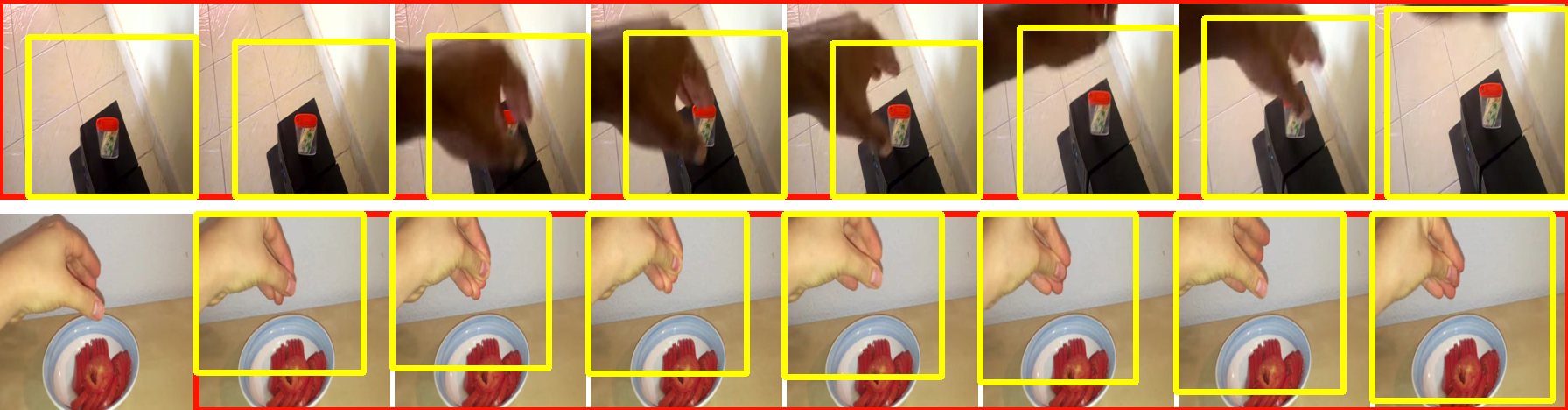}
    \includegraphics[width=0.45\linewidth]{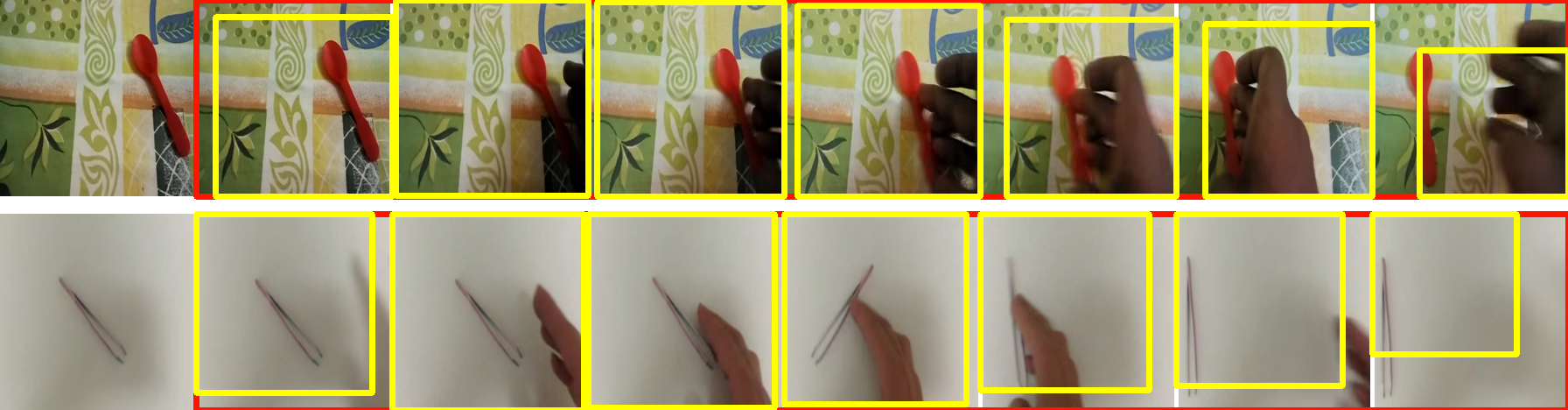}\\
    \includegraphics[width=0.45\linewidth]{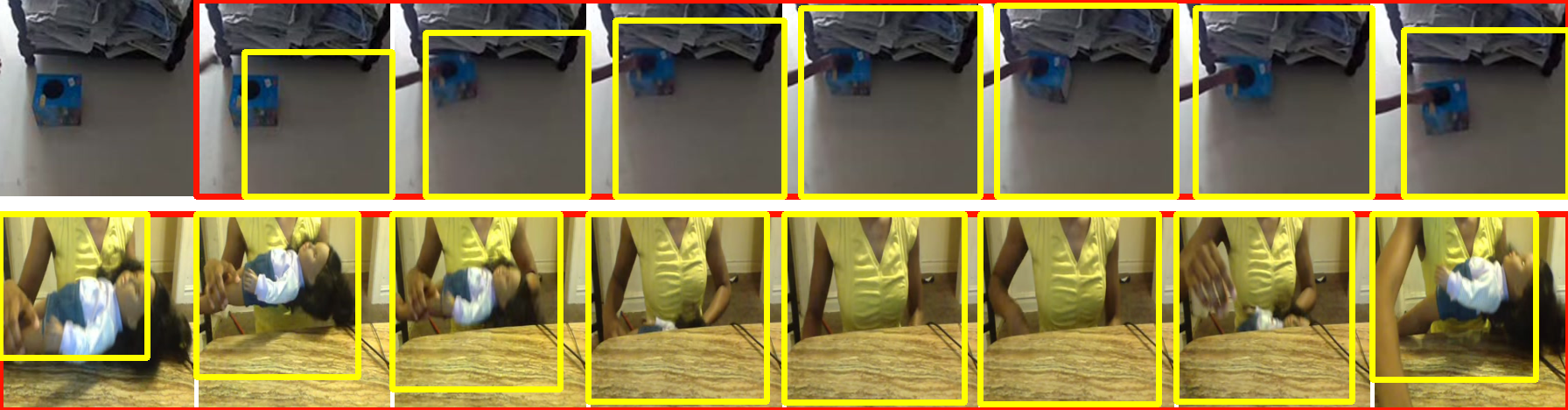}
    \includegraphics[width=0.45\linewidth]{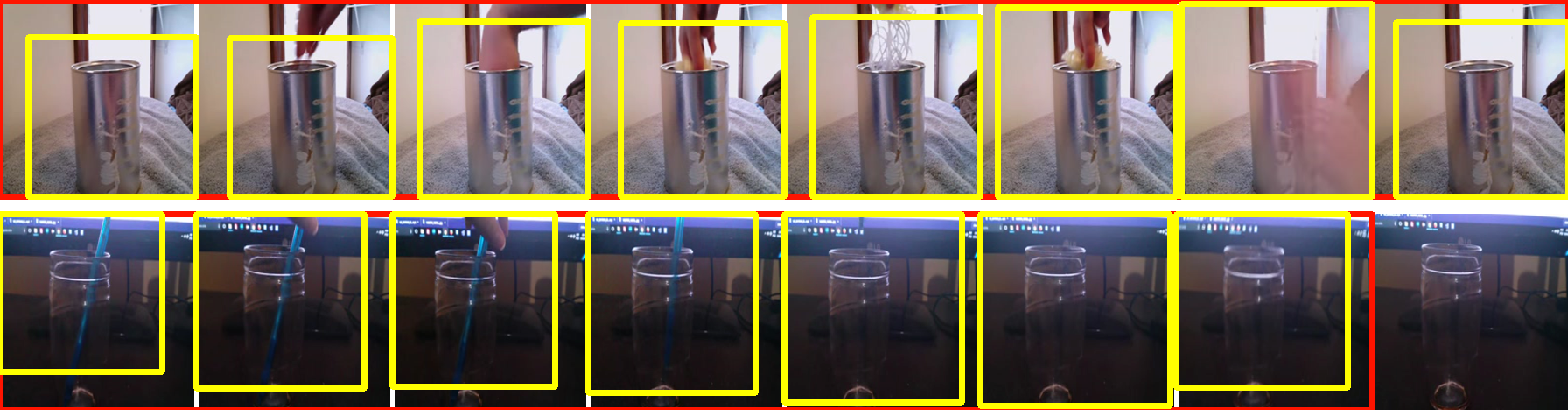}\\
    \caption{More visualisation on SSv2 dataset.  Action duration is highlighted with \textit{red} box and action-specific spatial region is highlighted with \textit{yellow} box.}
    \label{fig:my_label}
\end{figure*}

\subsection{Technical Details of Spatial Coordination}

\subsubsection{Offset Predictor} 

\begin{table}[ht]
\centering
\small
\begin{tabular}{l|l}
\hline
Layers & Kernel size \\ \hline
Conv3D+BN & k=3,pad=1,cout=128 \\
MaxPool3D+ReLU & k=(1,2,2) \\ \hline
Conv3D+BN & k=3,pad=1,cout=128 \\
MaxPool3D+ReLU & k=(1,2,2) \\ \hline
Spatial Global MaxPool3D & k=(1,+inf,+inf) \\ \hline
Conv1D+ReLU & k=1,cout=64 \\
Conv1D+Tanh & k=1,cout=2 \\ \hline
\end{tabular}
\caption{The structure of offset predictor $S$. k means kernel size and we use pytorch style dimension order (BCTHW).}
\label{tab:offset-pred}
\end{table}

We describe the detailed structure of proposed offset predictor in Tab.~\ref{tab:offset-pred}.

\subsubsection{Mask Generation}

Given predicted offset $o=[o_x,o_y] \in \mathbb{R}^{1\times 2}$, we generate masks $m_x$ and $m_y$ for $x$ and $y$ coordinates respectively. Then we can obtain the 2D offset mask $I_o = m_x \times m_y$. Specifically, $m_x$ is calculated by this piece-wise linear function:
\begin{equation}
\small
     m_x (o) =\begin{cases}
     max(0,1-\gamma(o_x-1-x))&, x \leq o_x-1  \\
     1                  &, |x-o_x|<1 \\
     max(0,1-\gamma(x-o_x-1))&, x \geq o_x+1 
     \end{cases}
\end{equation}
where $o_x$ is the offset in the $x$ dimension, the slope $\gamma=3$ in our case. $m_y$ is obtained in $y$ coordinate in the same way. The value of $\gamma$ ensures the width of the margin is about 1 pixel, which trades off between accurate result and differentiability.
$m_y$ is conducted in $y$ coordinate in the same way.

The perturbations added on the offset are 8-directional vectors, with amplitude decays for every 40 epochs. We average the masks generated with different perturbations to get the final one. This perturbation is disabled in testing.

\subsection{Prototype of Multiple Shots}

 In $1$-shot setting, the support prototype $\overline{p}_s = \overline{f}_s$. In the $k$-shot ($k>1$) FSL setting, the traditional strategy to generate \textit{prototype} of each class is simply averaging the features of all support samples. However, it ignores that the action misalignment also exists among support videos. Accordingly, we align them by applying Temporal Coordination (TC) over $k$ support features $\hat{f}_s$ (each of them is aligned by TTM firstly) to address this issue. 

Specifically, we randomly select a support feature from $k$ samples as the `\textit{reference}' for each class. Then, TC is applied align $k$-shot support samples to the `\textit{reference}': $\hat{f}_{s,i} = TC(\hat{f}_{s,i}, \hat{f}_{ref})$ ($i=1,\dots,k$). Finally, we average these aligned features to get the \textit{prototype}: $\hat{p}_s = \frac{1}{k}\sum_{i=1}^{k} \hat{f}_{s,i}$. 

Therefore, under multiple shot setting, query features $\hat{f}_q$ (aligned by TTM firstly) are then further aligned to these support prototypes $\hat{p}_s$ by action coordination module (ACM) to get the well-aligned query features $\overline{f}_q$ and support prototypes $\overline{p}_s$. Finally, given them, we can perform the classification according to Eq.~(7)-(8) (in our body text).

\subsection{Further Implementation Details}
We train our model using SGD optimizer, with initial learning rate of $1\times 10^{-3}$ and momentum of 0.9. For UCF101 and HMDB51, the learning rate decays by a factor of $0.5$ every 5 epochs. For SSv2, the learning rate decays by a factor of $0.8$ every 30 epochs. For kinetics, the learning rate decays by a factor of $0.9$ every 20 epochs without momentum in SGD. We train our model for 200 epochs for Kinetics, UCF101 and HMDB51, and 600 epochs for SSv2.

\subsection{More Visualization Results}

We show more visualization results in Fig.~\ref{fig:my_label}. 
\end{document}